# Intelligent Algorithm for Optimum Solutions Based on the Principles of Bat Sonar

Dr. Mohammed Ali Tawfeeq
Computer and Software Engineering Department
College of Engineering, Al-Mustansiriya University
Baghdad, Iraq
e-mail: drmatawfiq@yahoo.com

*Abstract*— **This paper presents a new intelligent algorithm that can solve the problems of finding the optimum solution in the state space among which the desired solution resides. The algorithm mimics the principles of bat sonar in finding its targets. The algorithm introduces three search approaches. The first search approach considers a single sonar unit (SSU) with a fixed beam length and a single starting point. In this approach, although the results converge toward the optimum fitness, it is not guaranteed to find the global optimum solution especially for complex problems; it is satisfied with finding "acceptably good" solutions to these problems. The second approach considers multisonar units (MSU) working in parallel in the same state space. Each unit has its own starting point and tries to find the optimum solution. In this approach the probability that the algorithm converges toward the optimum solution is significantly increased. It is found that this approach is suitable for complex functions and for problems of wide state space. In the third approach, a single sonar unit with a moment (SSM) is used in order to handle the problem of convergence toward a local optimum rather than a global optimum. The momentum term is added to the length of the transmitted beams. This will give the chance to find the best fitness in a wider range within the state space. The algorithm is also tested for the case in which there is more than one target value within the interval range such as trigonometric or periodic functions. The algorithm shows high performance in solving such problems. In this paper a comparison between the proposed algorithm and genetic algorithm (GA) has been made. It showed that both of the algorithms can catch approximately the optimum solutions for all of the testbed functions except for the function that has a local minimum, in which the proposed algorithm's result is much better than that of the GA algorithm. On the other hand, the comparison showed that the required execution time to obtain the optimum solution using the proposed algorithm is much less than that of the GA algorithm.**

## I. INTRODUCTION

The basic concept of any optimizing problem is to identify the alternative means of a given objective and then to select the alternative that accomplishes the objective in the most efficient manner, subject to constraints on the means. The problem can be represented mathematically as,

$$\text{Optimize } y = f(x_1, x_2, ..., x_n) \quad (1)$$

$$\text{Subject to } g_i(x_1, x_2, ..., x_n) = \begin{Bmatrix} \leq \\ = \\ \geq \end{Bmatrix} b_j \quad j=1, 2, ..., m \quad (2)$$

Equation (1) is the objective function and (2) constitutes the set of constraints imposed on the solution. The $x_i (i = 1,2,..., n)$ represent the set of decision variables, and $y=f(x_1, x_2, ..., x_n)$ is the objective function expressed in terms of these decision variables. Depending on the nature of the problem, the term optimize means either maximize or minimize the value of real function by systematically choosing input values from within an allowed set and computing the value of the function.

In general, optimization can be defined as the process of finding a best optimal solution for the problem under consideration.

Today, optimization comprises a wide variety of techniques. These techniques can be found in several literatures. Evolutionary computing may be the most prominent one in this field. In the 1950s and the 1960s several computer scientists independently studied evolutionary systems with the idea that evolution could be used as an optimization tool for engineering problems. The idea in all these systems was to evolve a population of candidate solutions to a given problem, using operators inspired by natural genetic variation and natural selection [1]. In 1975, Holland described how to apply the principles of natural evolution to optimization problems and built the first genetic algorithms (GA) [2]. In the last several years there have been widespread interaction among researchers studying various evolutionary computation methods, and the boundaries between GAs, evolution strategies, evolutionary programming, and other evolutionary approaches have broken down to some extent. These techniques are being increasingly widely applied to a variety of problems, ranging from practical applications in industry and commerce to leading-edge scientific research [3].

Particle swarm optimization (PSO) is another technique that optimizes a problem by iteratively trying to improve a candidate solution with regard to a given measure of quality. PSO is a form of swarm intelligence and is inspired by bird flocks, fish schooling and swarm of insects [2]. It is used as a heuristic search method for the exploration of solution spaces of complex optimization problems. Development on the PSO technique over the last decade has been made by different researchers. The heuristic in PSO suffers from relatively long execution times as the update step needs to be repeated many thousands of iterations to converge the swarm on the global optimum. Soudan, B. and Saad, M. [4] explored two dynamic population size improvements for classical PSO with the aim of reducing execution time. The most attractive features of PSO are its algorithmic simplicity and fast convergence. However, PSO tends to suffer from premature convergence when applied to strongly multimodal optimization problems. Lu H., et al. [5] proposed a method of incorporating a real-valued mutation (RVM) operator into the PSO algorithms, aimed at enhancing global search capability. The PSO contains many control parameters. These parameters cause the performance of the searching ability to be significantly alternated. In order to analyze the dynamics of such PSO system rigorously, Tsujimoto, T. et al. [6] proposed a canonical deterministic PSO system which does not contain any stochastic factors, and its coordinate of the phase space is normalized. The found global best information influences the dynamics. They regarded this situation as the full-connection state. The authors try to clarify the effective parameters on the CD-PSO performance. Feng Chen, et al. [7] proposed an improved PSO by incorporating the sigmoid function into the velocity update equation of PSO to tackle some drawbacks of PSO in order to obtain better global optimization result and faster convergence speed.

PSO shares many common points with GA. Both algorithms start with a group of a randomly generated population; both have fitness values to evaluate the population, both update the population and search for the optimum with random techniques. However, unlike GA, PSO has no evolution operators such as crossover and mutation. On the other hand, it is important to mention in this introduction that GAs and PSO do not guarantee success [2], and some times are not guaranteed to find the global optimum solution to a problem. They are satisfied with finding acceptably solutions to the problem.

This paper introduces a new intelligent algorithm. The proposed algorithm is a problem solving technique that uses the principles of bat sonar as its model in searching the approximate optimum solution for the problems. The algorithm introduces three search approaches, a single search unit, multisearch units, and a single search unit with a momentum. Each of these approaches can approximately find the optimum solution in solving the required problem with a reasonable efficiency depending on the complexity of the problem and the number of optimum points that exist in the problem.

This paper is organized as follows; the next section describes the main proposed algorithm. Section 3 introduces more efficient search approaches. Section 4 contains the experimental results, while, section 5 presents the conclusion of this work.

## II. MAIN ALGORITHM

The sonar of a bat is an active echolocation system. In addition to providing information about how far away a target is, bat sonar conveys information about the relative velocity of the target, the size of various features of the target, and the azimuth and elevation of the target [8].

In order to find its prey the bat may sit on a perch or fly around using its sonar signals. Some type of bats are considered as a 'high duty cycle' bat since it produces signals 80% of the time that it spends echolocating [9]. When a bat begins to echolocate it usually produces short millisecond long pulses of sonar, and listens to the returning echoes. If prey is detected by the bat, it will generally fly toward the source of the echo. The bat appears to be an amazing signal processing machine that has an accuracy of 99%. The way in which the bat can measure the distance and the size of its prey is as shown in Fig. 1 [10].

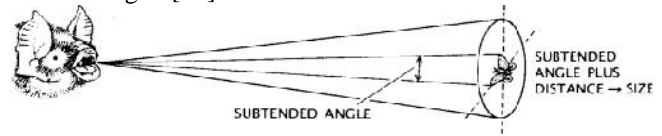

Fig. 1. Sonar signal of a bat

The proposed algorithm search for optimum solutions in problems depends mainly on these principles. In this algorithm, each and every point in the search space represents one possible solution. The sonar in this algorithm transmits several signals in different directions starting from a proposed starting point. Each transmitted signal contains a batch of N beams. These beams are of fixed length. The returned values (value of the fitness function at the end point of each beam) are checked with each other and compared with the starting point to determine the optimum one. If optimum point is detected, the sonar unit flies toward this point exchanging its starting point with the new one, then starts to transmit signals again from this point in different directions searching for better optimum solution. Otherwise, the sonar unit stays in its original starting point and retransmits signals in other directions. This process is repeated until the algorithm finds the best optimum solution.

Fig. 2 illustrates a sample on how the proposed algorithm searches for the optimum point. In this figure the sonar unit transmits beams of signals starting from point P1. The returned signals find better solution in P2. This causes the sonar to fly toward P2. This process is continued with P3, then with P4.

In this example, it is assumed that the entire returned signals to P4 are not fitter than P4, thus the algorithm considers P4 to be the optimum point.

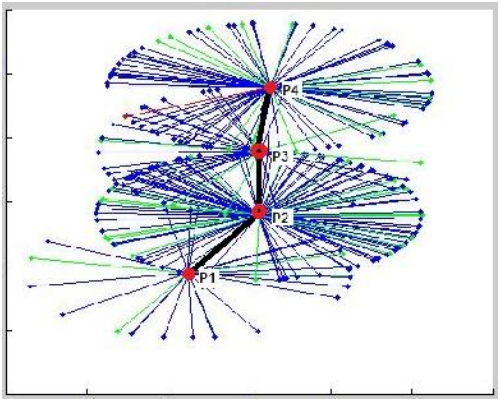

Fig. 2. Search process for optimum solution

The fitness function considered in the proposed algorithm, is the evaluation function that is used to determine the solution. This function can be n-dimensional. The optimal solution is the one with the best fitness function.

The main proposed algorithm in this paper considers a single sonar unit (SSU) flying in the state space searching for the optimum solution. This scenario represents the first search approach introduced in this work. The details of the algorithm are as follows:

***Step 1.*** Initialize the following main parameters:
- Solution range: min, max values of the search space variables.
- Beam length $L$: random value not exceeding half the solution range:

$$L \leq Rand * Solution\_range/2$$

- Number of beams $N$: Small integer random value representing the number of beams in each single transmitted signal.
- Starting point $pos_s$: any point in the search space selected randomly.
- Angle between beams: one of two techniques are assumed to be used in this algorithm. The first one is to randomly select a small fixed value between any two successive beams, while the other technique is to randomly select a different angle $\alpha_i$ between any two successive beams, where (i=1, ... , N-1). We called these two techniques "Fixed" and "Rand" respectively.

The above mentioned parameters are showed in Fig. 3.

***Step 2.*** Evaluate the fitness function at the start point $F_s$.

***Step 3.*** While stopping condition is false, do Steps 4-7.

***Step 4.*** Select random value representing the main beam direction $\alpha_m$ starting from $pos_s$.

***Step 5.*** Transmit $N$ beams starting from $pos_s$ with main beam direction of $\alpha_m$ and angle between any two successive beams.

***Step 6.*** Determine the coordinates of the remote end point $pos_i$ for each transmitted beam ($i=1,...,N$), then evaluate the fitness function $F_i$ at these ends. As an example, in a three dimension state space:

$$x_i = x_{pos_s} + L\cos(\alpha_m + (i-1)\alpha) \quad (3)$$

$$y_i = y_{pos_s} + L\sin(\alpha_m + (i-1)\alpha) \quad (4)$$

$$pos_i = [x_i, y_i]$$

$$F_i = f(x, y)$$

***Step 7.*** Compare the fitness values;
If $F_s$ is the optimum value (i.e., for maximizing $F_s \geq F_i$, and for minimizing $F_s \leq F_i$) then go to step 3
Otherwise:
Replace the coordinates of $pos_s$ with the coordinates of the optimum point of $F_i$ and replace $F_s$ with the optimum $F_i$:
$pos_s = pos_i$ of optimum $F_i$
$F_s$ = optimum $F_i$,
then go to step 3

***Step 8.*** Test for stopping condition: The algorithm can be terminated according to following stopping criteria:
- A fixed number of iterations have occurred.
- All solutions converge to the same value and no improvements in the fitness value are found.

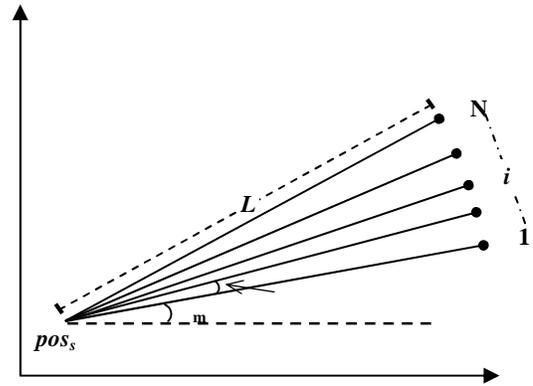

Fig. 3. Single batch of beams contained in a single transmitted signal

The algorithm is a kind of a parallel search; this comes from the fact that the technique used here is to check for several solutions at once. Over iterations, selection for best fitness leaves out bad solutions and gets the best in each step. Thus, the proposed algorithm tries to converge to optimal solutions.

In SSU, although the results converge toward the minimum or maximum fitness, it is not guaranteed to obtain the global optimum solution, especially in complex problems with wide state space. This leads to develop more efficient search approaches.

### III. MORE EFFICIENT SEARCH

This paper introduces two other more efficient search approaches, in which, the first one uses multisonar search units, while the other one adds a momentum term to the beams length. The backbone of these two algorithms is the main algorithm of SSU approach mentioned previously.

*A. Multisonar Search Units (MSU)*

This approach considers multisonar units (*m*) search for the optimum solution/s at the same moment. Each sonar unit has its own starting point. These units are working in parallel in the same search space. For an example Fig. 4 shows an MSU with three sonar units. This approach can be used in solving more complex, large search space problems, and in problems that have several optimum values. Because of the parallelism nature of this approach, MSU can reduce the execution time needed to find the optimum solution considerably especially in problems with large state space.

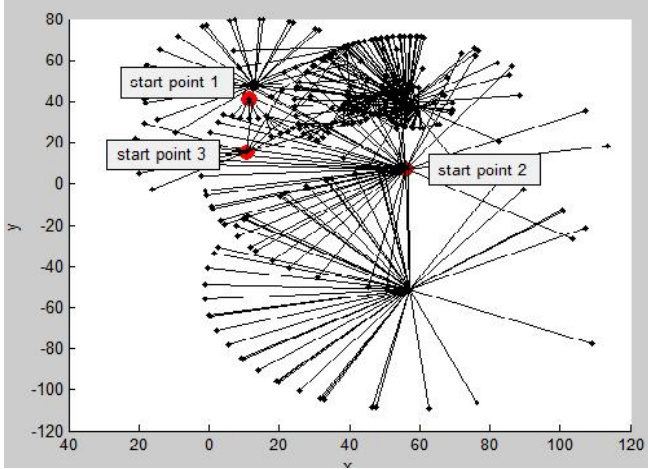

Fig. 4. MSU with three sonar units

*B. Single Sonar Units with A Momentum (SSM)*

Sometimes the solution found by SSU is not guaranteed to be the global optimum. This mainly comes from the nature of the problem and its state space, or due to the random selection of the initial parameters especially for the beam length. In such cases, the selected length of the transmitted beam, whatever the direction is, is either very long or very short so that it can not exceed to the area in which the solution is a global minima or maximum. SSM introduces a momentum term μ in order to reduce the problem of convergence toward a local optimum. In this approach, when the sonar unit converges toward an optimum solution, this solution will be checked again to be assured that it is not a local optimum. The proposed technique used here is to add a momentum term to the length of the transmitted beams. Using a momentum gives the chance to search for optimum solution in a wider range within the search space. The value of the momentum is considered to be within the range 0<μ<1. The new beam length then becomes:

$$L_{new} = L_{old}(1 \pm \mu) \quad (5)$$

## IV. EXPERIMENTAL RESULTS

Different types of fitness functions are used to test and evaluate the proposed algorithm with its three search approaches. In these tests, some of the initial parameters are considered to be the same. This will give the chance to make a worthy and meaningful comparison between the performances and the efficiencies of the three approaches. These parameters are:

Number of beams in each transmitted signal N=5
Angle between any two successive beams for Fixed technique = π/12
Maximum number of iterations in each run = 100
Number of runs (epochs) = 500

The performance of each approach is considered to be the degree on how much the obtained solution meets the goal. Where the goal is assumed here as the value that is equal or approximately equal to the optimum solution. Thus, for each solved function, the overall performance η for the used approach is determined as,

$$\eta = S_g / M \times 100\% \quad (6)$$

Where, $S_g$ is the number of the obtained solutions greater than or equal to the goal, and *M* is the total number of epochs. In this work two goal values are considered. The first one is assumed to be greater than 97.5% of the optimum solution, while the second one is greater than 96% of the optimum solution. The corresponding calculated performances for these goals are named $\eta_1$ and $\eta_2$ respectively.

And the overall efficiency γ is calculated as

γ = *AverageObtainedFitness/OptimumFitness* ×100%

$$\gamma = Avg(F_{obt}) / F_{opt} \times 100\% \quad (7)$$

The average Euclidean distance $\|E_d\|$ between the obtained solutions and the optimum one are calculated as follows:

$$\|E_d\| = \frac{1}{M}\sqrt{\sum_{i=1}^{M}(F_{opt_i} - F_{obt_i})^2} \quad (8)$$

Where,
$F_{opt}$ is the fitness of the optimum solution,
$F_{obt}$ is the fitness of the obtained solution using the proposed algorithm.

The used fitness functions and their tests results are as follows:

*A.* The first used function is a third order polynomial with a single variable. This function is described as

$$F_i = f_1(x) = x^3 - 5x^2 - 20x \quad (9)$$

It is required to find the maximum value of this function within an assumed range of -6<x<6. The algebraic calculation shows that the maximum value of this function is about 15.4564 at x= -1.4064.

In this work, the SSU approach is tested to find the optimum fitness value. Fig. 5 shows one epoch as an example on how SSU search for the best fitness. The obtained results for the 500 epochs are summarized in table 1.

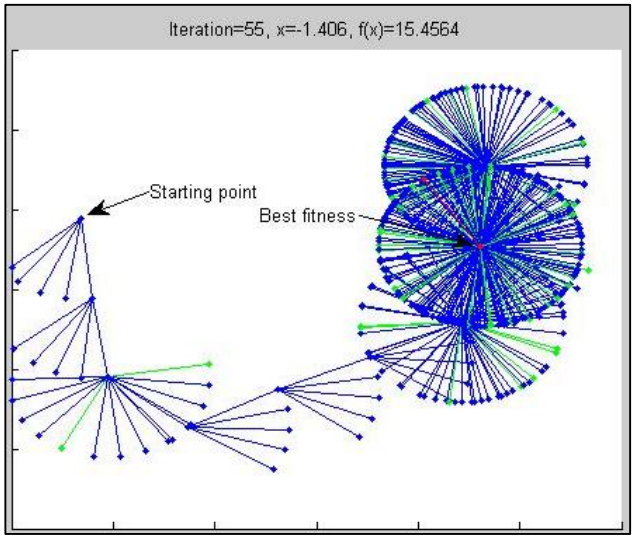

Fig. 5. SSU search for optimum fitness

TABLE 1
SUMMERY OF THE OBTAINED RESULTS OF $f_1$ USING SSU

| Max obtained fitness | 15.4564 |
|---|---|
| x | -1.4064 |
| $Avg(F_{obt})$ | 15.4294 |
| Avg. no. of iterations | 50.2 |
| y | 99.8% |
| ...1 | 99.8% |
| ...2 | 100% |
| $\|E_d\|$ | 0.003 |

The results of this test show that the best obtained fitness matches the maximum calculated fitness with high efficiency. Fig. 6 shows the obtained solutions, in which each dot in this figure represents an obtained fitness. In this test most of the obtained values are very close to each other, and some of them are equal, thus they appear in this figure as overlapped dots.

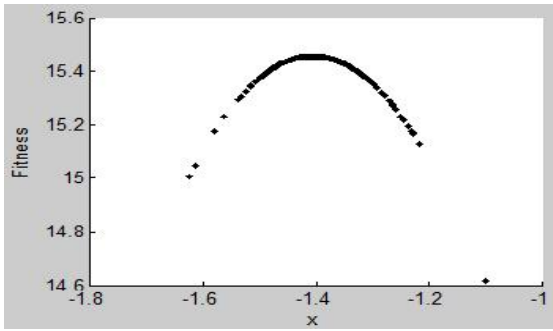

Fig. 6. Obtained solution of $f_1$ using SSU

Although the test results of using SSU are good, but it is not guaranteed to find the global optimum solution for complex problems with wide state space.

*B.* The second used function is a fifth order polynomial of a single variable described by (10). It is also required to find the max value of this function within the range of -6<x<6.

$$F_i = f_2(x) = x^5 - 10x^4 - 5.2x^3 - 12x^2 + 5.5x \quad (10)$$

The maximum calculated value of $f_2$ is 0.5635 at x= 0.1932. In this test, we first examined the following approaches: SSU, MSU with two sonar units, and MSU with three sonar units. Each approach tested using Fixed and Rand techniques respectively. The obtained results are summarized in table 2.

TABLE 2
SUMMERY OF THE OBTAINED RESULTS OF $f_2$ USING SSU AND MSU

|  | SSU | | MSU (2 units) | | MSU (3 units) | |
|---|---|---|---|---|---|---|
|  | Fixed | Rand | Fixed | Rand | Fixed | Rand |
| Max($F_{obt}$) | 0.5635 | 0.5635 | 0.5635 | 0.5635 | 0.5635 | 0.5635 |
| x | 0.1932 | 0.1932 | 0.1932 | 0.1932 | 0.1932 | 0.1932 |
| $Avg (F_{obt})$ | 0.5560 | 0.5536 | 0.5587 | 0.5585 | 0.5625 | 0.5628 |
| Avg. no. of iterations | 53.3 | 54.7 | 51 | 53.7 | 53.7 | 56.7 |
| y | 98.67 | 98.2% | 99.1% | 99.1% | 99.82% | 99.87% |
| ...1 | 85.2% | 81.2 | 91% | 93.8% | 99.6% | 99.8% |
| ...2 | 90.8% | 90.4% | 95.6% | 95.8% | 99.8% | 100% |
| $\|E_d\|$ | $7\times10^{-4}$ | $1\times10^{-3}$ | $5\times10^{-4}$ | $1.2\times10^{-3}$ | $1.2\times10^{-4}$ | $7.9\times10^{-5}$ |

The obtained solutions of $f_2$ using MSU approach with three sonar units and Rand technique are shown in the three parts of Fig. 7, in which, each part represents one search unit.

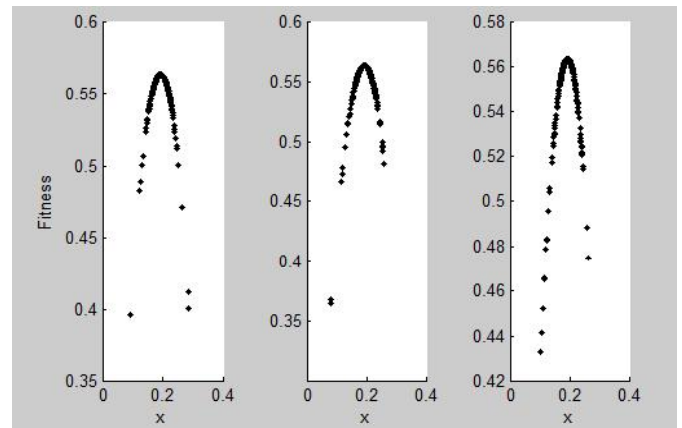

Fig. 7. Obtained solution of $f_2$ using MSU with three sonar units

The test of the SSU approach shows low performance. This performance can be improved by using the SSM approach, in which a momentum term $\mu$ is added to the length of the transmitted beams. By applying a momentum $\mu=0.9$ and solving for $f_2$ using both Fixed$_\theta$ and Rand$_\theta$ techniques, the performance is significantly increased to be 100%, with much better Euclidean distance as shown in table 3.

TABLE 3
RESULTS OF $f_2$ USING SSU AND SSM

|  | SSU | | SSM | |
|---|---|---|---|---|
|  | Fixed | Rand | Fixed | Rand |
| Max($F_{obt}$) | 0.5635 | 0.5635 | 0.5635 | 0.5635 |
| x | 0.1932 | 0.1932 | 0.1932 | 0.1932 |
| $Avg (F_{obt})$ | 0.5560 | 0.5536 | 0.5634 | 0.5634 |
| Avg. no. of iterations | 53.3 | 54.7 | SSU+ 46.8 | SSU+ 47 |
| y | 98.67 | 98.2% | 99.98% | 99.98% |
| ...1 | 85.2% | 81.2 | 100% | 100% |
| ...2 | 90.8% | 90.4% | 100% | 100% |
| $\|E_d\|$ | $7.0\times10^{-4}$ | $1\times10^{-3}$ | $4.2\times10^{-6}$ | $7.9\times10^{-6}$ |

The obtained fitness of $f_2$ using SSU is shown in part (a) of Fig. 8, while part (b) shows the results of using SSM, in which, the obtained fitness are constrained to be very close to the optimum solution.

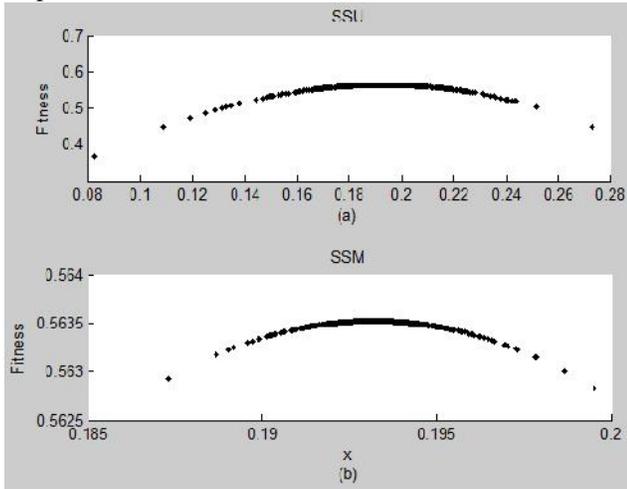

Fig. 8. Obtained fitness of $f_2$ (a) using SSU (b) using SSM

*C*. The third tested function is a polynomial with two variables described by (11) and shown in Fig. 9.

$$F_i = f_3(x, y) = x^3 - 5x^2 - 2.04y^2 + 4y \quad (11)$$

The ranges of the solution space for this function are taken to be -3<x<3, and -3<y<3. The maximum calculated value of $f_3$ is 1.9608 at x=0 and y=0.9809.

The three proposed approaches are tested for the convergence toward the maximum fitness of $f_3$. The obtained results are summarized in table 4. In this test, although the efficiency of the SSU approach is good but its performance is not accepted in solving such a problem. The alternative is to use either the SSM approach which gives much better performance, or to use the MSU approach with not more than three search units, in which the performance is increased to be about 100% with high overall efficiency. The distribution of the x, y values for the obtained fitness using SSU and SSM are shown in Fig. 10, while Fig. 11 shows this distribution when using the MSU approach.

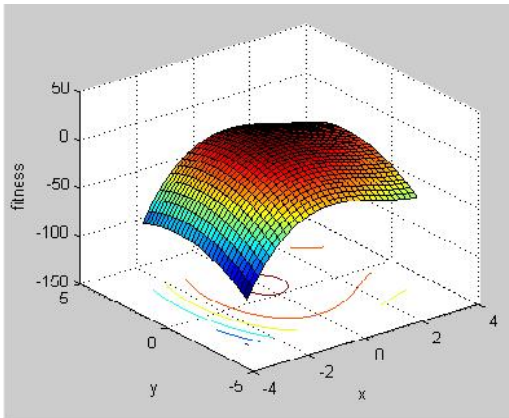

Fig. 9. Polynomial function ($f_3$) with two variables

TABLE 4
TEST RESULTS OF SOLVING $f_3$ USING SSU, SSM AND MSU

|  | SSU | SSM | MSU (3 units) |
|---|---|---|---|
|  | Rand | Rand | Rand |
| Max($F_{obt}$) | 1.9607 | 1.9608 | 1.9608 |
| x | -0.0019 | 0 | 0 |
| y | 0.9869 | 0.9826 | 0.9826 |
| Avg ($F_{obt}$) | 1.887 | 1.9264 | 1.9505 |
| Avg. no. of iterations | 33 | SSU+ 11 | 55 |
| y | 96.2% | 98.2% | 99.48% |
| ...1 | 30% | 85% | 99.8% |
| ...2 | 60% | 98.2% | 100% |
| $||E_d||$ | 3.7×10⁻³ | 1.7×10⁻³ | 6×10⁻⁴ |

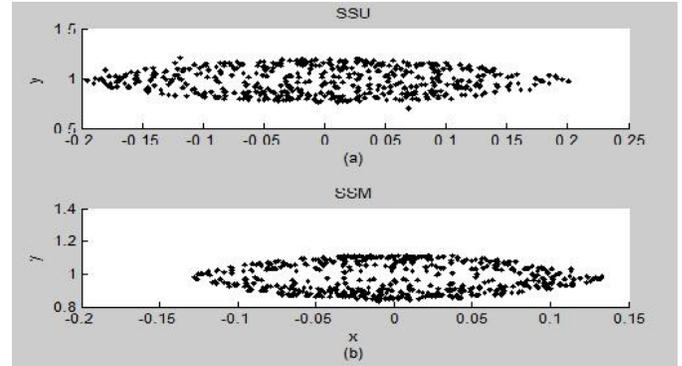

Fig. 10. Distribution of x, y values of $f_3$ fitness (a) using SSU, (b) using SSM

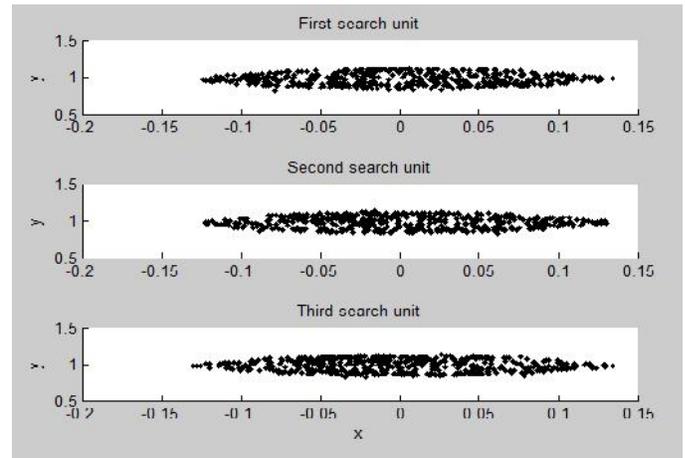

Fig. 11. Distribution of x, y values of $f_3$ fitness using MSU approach with three search units

*D*. The fourth case tests an exponential function with two variables. This function is described by (12). The ranges of the solution space are taken to be between -2 to +2 for both x and y as shown in Fig. 12.

$$F_i = f_4(x, y) = xe^{(-x^2 - y^2)} \quad (12)$$

The maximum calculated value for $f_4$ is 0.4289 at x=0.7072 and y=0. The obtained results of using SSU, SSM, and MSU (with three points) are contained in table 5. The three approaches converged toward the optimum point with different performances. The performance of using SSU in solving functions like $f_4$ is very low. Rather than the use of this approach its better to apply either the SSM approach or the MSU approach in which both have a performance of 100% with less Euclidean distance. The distribution of the x, y values for the obtained solutions using the above mentioned approaches are shown in Fig. 13 and Fig. 14.

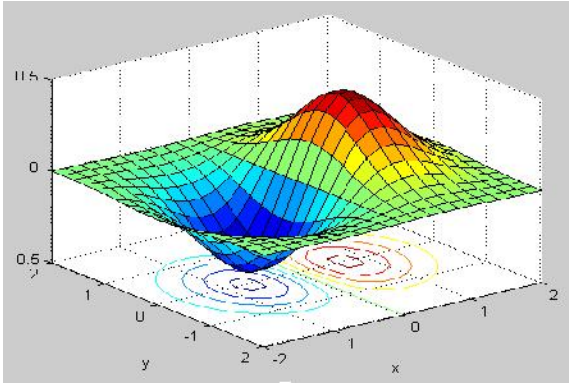

Fig. 12. Exponential function ($f_4$) with two variables

TABLE 5
TEST RESULTS OF $f_4$ USING SSU, SSM AND MSU

|  | SSU | SSM | MSU (3 units) |
|---|---|---|---|
|  | Rand | Rand | Rand |
| Max($F_{obt}$) | 0.4289 | 0.4289 | 0.4289 |
| x | 0.7086 | 0.7086 | 0.7083 |
| y | -0.0071 | -0.0071 | -0.0006 |
| Avg ($F_{obt}$) | 0.412 | 0.4283 | 0.4277 |
| Avg. no. of iterations | 21 | SSU+ 29.2 | 62 |
| η | 96 % | 99.97% | 99.7% |
| ...1 | 30% | 100% | 100% |
| ...2 | 53% | 100% | 100% |
| //$E_d$// | 8.7×10⁻⁴ | 5.2×10⁻⁶ | 8.1×10⁻⁵ |

*E.* The last test in this work considers a function with several optimum points and checks the ability of the proposed algorithm to converge towards these points. A trigonometric or a periodic function is a good example for such a case, in which these types of functions repeat their values in regular intervals or periods. The selected function for this test is:

$$F_i = f_5(x) = \sin(2x) - \cos(x) \quad (13)$$

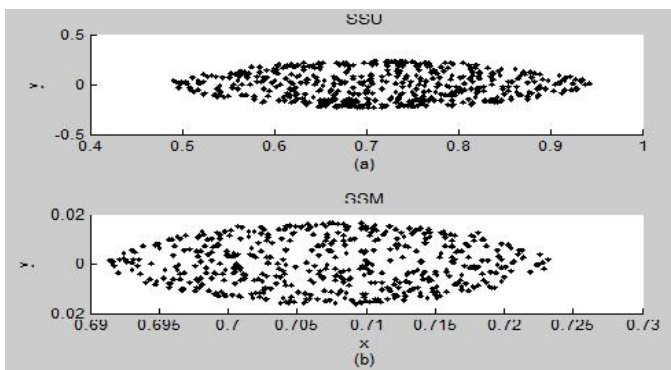

Fig. 13. Distribution of x, y values of $f_4$ fitness using SSU and SSM approaches

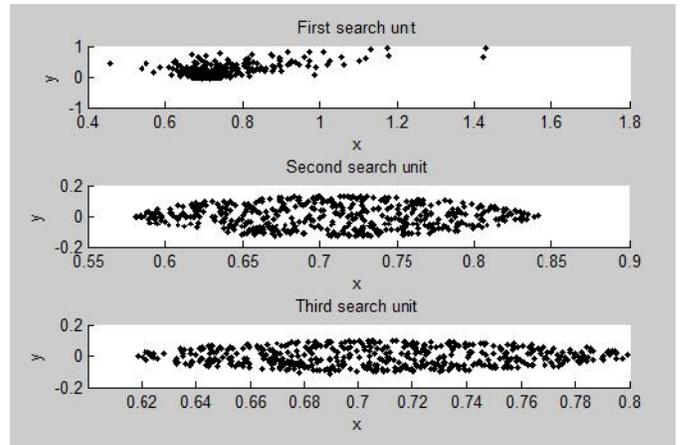

Fig. 14. Distribution of x, y values of $f_4$ fitness using MSU approach with three search units

The solution range is assumed to be between -2π to 2π. Within this interval, the function $f_5$ has two optimum values of about 1.76017 at x=-2.5067 and x=3.7765 as shown in Fig. 15.

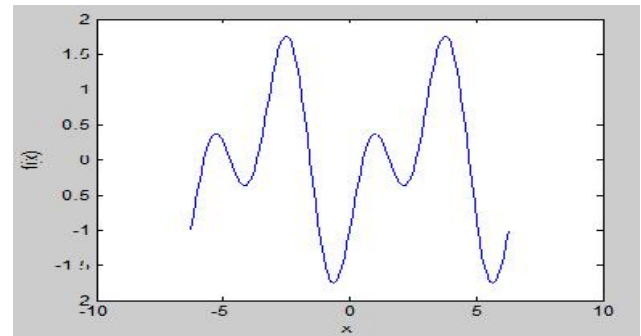

Fig. 15. Periodic function ($f_5$) with two global optimum values

The tested approach is the MSU with two sonar units. As mentioned before, in this approach, the search units are working in parallel in the same state space. The obtained results showed that, either both of the search units converged towards the same optimum point, or each unit converged toward a different optimum point, but in general, the algorithm observed the two global optimum points in high performance levels. The test results are as shown in table 6. It is found that the overall average fitness is very close to the optimum value (η = 99.68%) and a performance between 98.6% and 100% with acceptable Euclidean distance. The obtained fitness for this test is shown in Fig. 16.

TABLE 6
OBTAINED RESULTS OF $f_5$ USING MSU WITH TWO UNITS

| Max($F_{obt}$) | 1.76017 |
|---|---|
| $x_1$ | -2.5070 |
| $x_2$ | 3.7765 |
| Avg ($F_{obt}$) | 1.7544 |
| Avg. no. of iterations | 44 |
| η | 99.68% |
| ...1 | 98.6% |
| ...2 | 100% |
| //$E_d$// | 5.5×10⁻⁵ |

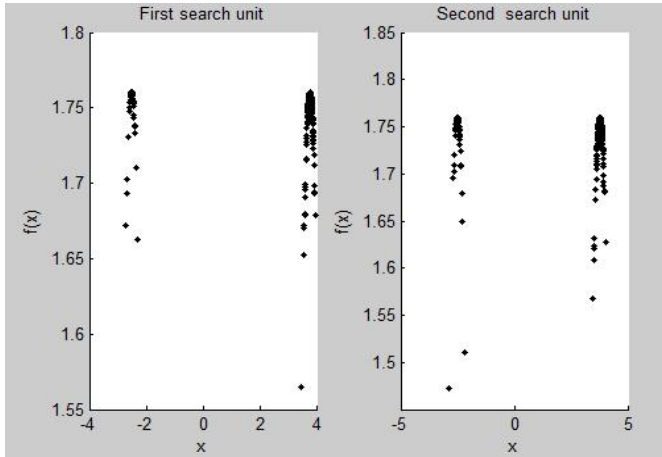

Fig. 16. Obtained fitness of $f_5$ using MSU with two search units

In order to evaluate the proposed algorithm, a comparison with genetic algorithm has been made, in which, the obtained fitness and the execution time for both of the algorithms are tested using the above mentioned five functions. The used platform for the two algorithms is "MATLAB® R2010b 32-bit (win32)". The results of the comparison are as declared in table 7.

In this comparison, the obtained fitnesses for the first four testbed functions are approximately the same in both of the algorithms. In function $f_5$, shown in Fig. 15, it is clear that there are two global and two local optimum points within the considered range space. The comparison showed that the obtained result of solving f5 using the MSS (with 2 sonar units) is much better than the result obtained by using GA. In which, the MMS algorithm observed the two global points as its best fitness, while the obtained fitness using GA algorithm is one of the local optimum.

TABLE 7
COMPARISON BETWEEN THE PROPOSED ALGORITHM AND GA ALGORITHM

| Function | Algorithm | $F_{obt}$ | x | y | Execution time (msec) |
|---|---|---|---|---|---|
| f1 | SSU | 15.4564 | -1.40640 | -- | 1.7 |
| f1 | GA | 15.4562 | -1.40180 | -- | 116 |
| f2 | SSM | 0.56350 | 0.19320 | -- | 3.2 |
| f2 | GA | 0.56350 | 0.19320 | -- | 127 |
| f3 | SSM | 1.96080 | 0.00000 | 0.98190 | 3 |
| f3 | GA | 1.96080 | 0.00000 | 0.98037 | 119 |
| f4 | SSM | 0.42890 | 0.70860 | -0.0071 | 2.7 |
| f4 | GA | 0.42890 | 0.70700 | 0.00020 | 123 |
| f5 | SSM | 1.76017 / 1.76017 | -2.50674 / 3.77650 | -- | 2.6 |
| f5 | MSU with 2 units | 1.76017 / 1.76017 | -2.50674 / 3.77650 | -- | 3.5 |
| f5 | GA | 0.36900 | 1.00300 | -- | 112 |

The other parameter that takes place in this comparison is the execution time required to solve each function. This comparison based on the following considerations:

- In the proposed algorithm, a single epoch with number of iterations = 100 is considered in solving each function.
- In GA algorithm, the default setting of the MATLAB built in function "ga" is considered, in which the maximum allowed number of generations before the algorithm halts is = 100.

As it is seen from table 7, the calculated execution times using the proposed algorithm are much less than those of the GA algorithm.

## V. CONCLUSION

Intelligent algorithms are, in many cases, practical alternative techniques for solving a variety of challenging engineering problems. These techniques are, in general, attempts to mimic some of the processes taking place in natural evolution. This paper introduces a new intelligent algorithm with three search approaches depending mainly on the principles of how bat sonar can detect and capture its target. The first approach uses a single sonar unit in its search process. While the second one uses multisearch units working in parallel. This approach has been developed to reduce the execution times that are associated with the use of the first approach for finding near-optimal solutions in large search spaces and to find better solutions in larger problems. The third approach uses a single search unit with a momentum term added to the beam length. The three approaches are tested on different types of functions, such as; polynomial, exponential, and trigonometric or periodic functions. The search results show that the proposed algorithm approximately recognized all the optimum values with a reasonable efficiency and "acceptable to high" performance depending on the complexity of the problem and the number of optimum points that exist in the problem.

Although the initial values of the main parameters are selected randomly, some complex problems may need a heuristic signal to decrease the execution time and to find the optimum solution with high performance. This signal can mainly be used for the selection of the initial value of the beam length.

A comparison between the proposed algorithm of this paper and GA algorithm showed that the proposed algorithm is much better in solving problems having a local optimum. On the other hand, the required execution times for the proposed algorithm are much less than that of the GA algorithm for all the tested functions.

From the optimization point of view, the main advantage of the approaches introduced in this paper is that they do not have much mathematical requirements. All they need is an evaluation of the objective function. As a result, they can be easily applied to solving a wide class of scientific and engineering optimization problems.